\documentclass{article}

\usepackage{arxiv}

\usepackage[utf8]{inputenc} 
\usepackage[T1]{fontenc}    
\usepackage{hyperref}       
\usepackage{url}            
\usepackage{booktabs}       
\usepackage{amsfonts}       
\usepackage{nicefrac}       
\usepackage{microtype}      
\usepackage{lipsum}		
\usepackage{graphicx}
\usepackage{natbib}
\usepackage{doi}

\usepackage{algorithm}
\usepackage{algpseudocode}
\usepackage{amsmath}

\title{Combining Entropy and Matrix Nuclear Norm for Enhanced Evaluation of Large Language Models}

\author{ \href{https://orcid.org/0000-0002-4363-2177}{\includegraphics[scale=0.06]{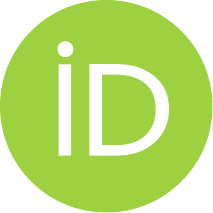}\hspace{1mm}James Vo}\thanks{Anh-Dung Vo} \\
	DocumentAI Team\\
	AGILESODA INC.\\
	Seoul, 06149, South Korea \\
	\texttt{anhdungitvn@agilesoda.ai} \\
}

\renewcommand{\shorttitle}{Combining Entropy and Matrix Nuclear Norm for Enhanced Evaluation of LLMs}

\hypersetup{
pdftitle={Combining Entropy and Matrix Nuclear Norm for Enhanced Evaluation of Language Models},
pdfsubject={q-bio.NC, q-bio.QM},
pdfauthor={David S.~Hippocampus, Elias D.~Striatum},
pdfkeywords={First keyword, Second keyword, More},
}

\begin{document}
\maketitle

\begin{abstract}

As large language models (LLMs) continue to advance, the need for precise and efficient evaluation metrics becomes more pressing. Traditional approaches, while informative, often face limitations in computational demands and interpretability. In this paper, we introduce a novel hybrid evaluation method that integrates two established techniques: entropy derived from covariance matrices and the Matrix Nuclear Norm (MNN). Our method begins by normalizing hidden states from LLMs, then computes the covariance matrix and MNN from these representations. We further calculate the entropy of the covariance matrix to capture uncertainty and redundancy in the model's outputs. By combining these metrics into a composite score, we offer a comprehensive evaluation framework that balances accuracy with computational efficiency. Additionally, our approach allows for flexibility in adjusting the weightings between entropy and MNN, tailoring the evaluation for different objectives. Through a series of experiments on various LLMs, we demonstrate the robustness and efficacy of our method, offering deeper insights into model performance. This work contributes to the ongoing development of LLM evaluation and opens avenues for future innovations in model assessment techniques.

\end{abstract}

\keywords{Large Language Models \and Matrix Nuclear Norm \and Entropy-based Evaluation \and Model Performance Assessment \and Representation Analysis}

\section{Introduction}

Large Language Models (LLMs) have revolutionized the field of natural language processing (NLP) by demonstrating unprecedented capabilities in understanding and generating human-like text. Models such as GPT-4, BERT, and their successors have not only achieved remarkable performance across a variety of tasks but have also extended their utility into multi-modal domains, encompassing vision, audio, and other data types. As these models continue to grow in size and complexity, evaluating their performance accurately and efficiently becomes increasingly critical.

Traditional evaluation metrics for LLMs, including perplexity, accuracy, and F1 scores, primarily focus on task-specific outcomes. While these metrics provide valuable insights into a model's ability to perform particular tasks, they often fall short in capturing the underlying representational dynamics and information compression capabilities of the models. Moreover, as LLMs scale, the computational demands of these conventional metrics can become prohibitive, necessitating the development of more sophisticated and scalable evaluation methodologies.

Recent advancements have introduced novel metrics that delve deeper into the internal workings of LLMs. One such approach is Diff-eRank \cite{wei2024differanknovelrankbasedmetric}, a rank-based metric grounded in information theory and geometric principles. Diff-eRank assesses LLMs by analyzing their hidden representations, offering a quantitative measure of how effectively these models eliminate redundant information during training. This method provides statistical insights into model generalization and ensures normalization of data, thereby enhancing evaluation robustness. However, it relies on specific distribution assumptions of singular values and presents challenges in interpreting entropy values relative to model performance.

Another promising metric is the MNN \cite{li2024largelanguagemodelevaluation}, which measures matrix complexity as the sum of its singular values. MNN serves as a convex approximation of matrix rank, capturing both predictive discriminability and diversity while significantly reducing computational complexity from \( O(n^3) \) to \( O(n^2) \). This efficiency makes MNN particularly suitable for large-scale models, achieving speeds up to 24 times faster than Matrix Entropy for models like CEREBRAS-GPT. Despite its advantages, MNN can be computationally intensive for very large matrices and may incur information loss depending on the selected top dimension.

While both Diff-eRank and MNN offer valuable perspectives on LLM evaluation, each has its inherent limitations. Diff-eRank provides deep statistical insights but may struggle with interpretability and distributional assumptions. On the other hand, MNN offers computational efficiency and a solid mathematical foundation but may face challenges related to computational intensity and potential information loss.

To address these challenges, this paper proposes a novel hybrid evaluation method that synergistically combines the strengths of both Diff-eRank and MNN. By integrating entropy derived from covariance matrices with the MNN, our approach aims to provide a more comprehensive and balanced assessment of LLM performance. This hybrid method not only leverages the statistical depth of entropy-based evaluations but also incorporates the computational efficiency and representational insights offered by MNN. The flexibility to tune the weights between entropy and MNN further allows for customization based on specific evaluation objectives and datasets.

The contributions of this study are threefold:
\begin{enumerate}
    \item \textbf{Hybrid Evaluation Metric}: Introduction of a composite score that integrates entropy and MNN, offering a nuanced evaluation framework for LLMs.
    \item \textbf{Computational Efficiency}: Demonstration of reduced computational complexity compared to traditional metrics, ensuring scalability to large models.
    \item \textbf{Empirical Validation}: Comprehensive experiments across various LLMs to validate the robustness and effectiveness of the proposed hybrid method.
\end{enumerate}

By advancing the evaluation methodologies for LLMs, this work not only enhances our ability to assess model performance more accurately but also contributes to the broader understanding of the intricate dynamics governing large-scale language models. The proposed hybrid method sets the stage for future explorations into innovative evaluation techniques, fostering the development of more efficient and interpretable NLP models.

\section{Methodology}

Our proposed evaluation framework combines two complementary metrics, Entropy from Covariance Matrices and MNN, to offer a robust and scalable hybrid approach for evaluating LLMs. These metrics capture different dimensions of the model’s internal representations, enabling a comprehensive evaluation across both geometric and information-theoretic perspectives. In the following sections, we systematically present each component of our methodology, from data preparation and normalization to final score computation, providing detailed mathematical formulations and justifications along the way.

\subsection{Data Normalization}

The first step in our methodology is normalizing the hidden states obtained from the LLMs. Hidden states represent the internal activations generated by the model as it processes sequences of tokens, typically forming high-dimensional vectors that encode both semantic and syntactic information. Since these vectors may vary significantly in scale between different models or datasets, normalization is essential for ensuring a fair and unbiased comparison across models.

We employ a two-step normalization process, combining mean-centering and L2-norm scaling to produce stable and comparable representations:

\begin{itemize}
    \item \textbf{Mean-Centering}: Each hidden state vector is centered by subtracting the mean of the hidden states across all tokens in the sequence. This eliminates any shift in the data and ensures that the hidden states have a zero mean. Mathematically, for a set of hidden states \( \{ \mathbf{h}_i \}_{i=1}^n \), where \( \mathbf{h}_i \in \mathbb{R}^d \) represents the hidden state of the \( i \)-th token and \( d \) is the dimension of the hidden states, the mean-centered vector \( \mathbf{h}_i' \) is computed as:
    \[
    \mathbf{h}_i' = \mathbf{h}_i - \bar{\mathbf{h}}, \quad \text{where} \quad \bar{\mathbf{h}} = \frac{1}{n} \sum_{i=1}^{n} \mathbf{h}_i.
    \]
    \item \textbf{L2-Norm Scaling}: After mean-centering, each hidden state vector is scaled by its L2-norm, ensuring that all vectors lie on the unit hypersphere. This step is critical for removing the influence of sequence length or model-specific scaling factors, and is formally expressed as:
    \[
    \mathbf{h}_i^{\text{norm}} = \frac{\mathbf{h}_i' }{\| \mathbf{h}_i' \|_2}, \quad \text{where} \quad \| \mathbf{h}_i' \|_2 = \sqrt{\sum_{j=1}^{d} \mathbf{h}_{ij}'^2}.
    \]
\end{itemize}

This two-stage normalization ensures that the hidden states are centered and normalized, providing a standardized input for the subsequent analysis and making the covariance matrix calculation more stable and comparable.

\subsection{Covariance Matrix Calculation}

Once the hidden states are normalized, we compute the covariance matrix of these representations. The covariance matrix is central to our analysis as it captures the pairwise relationships between the different dimensions of the hidden state vectors, offering insights into how the model encodes information and how the dimensions interact.

Let \( \mathbf{H} \in \mathbb{R}^{n \times d} \) be the matrix of normalized hidden states, where \( n \) represents the number of tokens in the sequence, and \( d \) is the dimensionality of the hidden states. The covariance matrix \( \Sigma \in \mathbb{R}^{d \times d} \) is computed as:
\[
\Sigma = \frac{1}{n} \mathbf{H}^T \mathbf{H}.
\]
Here, \( \mathbf{H}^T \mathbf{H} \) represents the inner product of the hidden state matrix with itself, and dividing by \( n \) ensures that the covariance matrix is properly scaled by the number of tokens. This matrix encapsulates how different dimensions of the hidden states co-vary, providing valuable information about the model's internal representation structure.

\subsection{Entropy Calculation}

With the covariance matrix \( \Sigma \) in hand, we proceed to calculate the entropy of the covariance matrix. Entropy, a key concept in information theory, measures the uncertainty or randomness in a system, and in our context, it quantifies the amount of information encoded by the model’s hidden states.

To compute entropy, we first perform a Singular Value Decomposition (SVD) on the covariance matrix \( \Sigma \), yielding singular values \( \sigma_1, \sigma_2, \dots, \sigma_d \). These singular values correspond to the variances along the principal components of the hidden states. Once the singular values are obtained, we normalize them as:
\[
\tilde{\sigma}_i = \frac{\sigma_i}{\sum_{j=1}^{d} \sigma_j}, \quad \text{for each} \ i = 1, \dots, d,
\]
which ensures that they form a valid probability distribution. The entropy of the covariance matrix is then computed using the formula:
\[
\text{Entropy}(\Sigma) = -\sum_{i=1}^{d} \tilde{\sigma}_i \log(\tilde{\sigma}_i),
\]
where the logarithm is typically taken in base 2, making the entropy value interpretable in bits. This entropy value reflects the degree of diversity in the hidden states, with higher entropy values indicating a more complex, diverse representation, while lower entropy values suggest more redundant or compressed representations.

\subsection{MNN Calculation}

In addition to entropy, we compute the MNN, which serves as a convex approximation of the rank of the hidden state matrix. The nuclear norm is particularly useful in our context as it provides a measure of the spread of the singular values, indicating how the hidden states occupy the representational space.

The nuclear norm is defined as the sum of the singular values of the matrix. Given the singular values \( \sigma_1, \sigma_2, \dots, \sigma_d \) obtained from the SVD of the covariance matrix, the MNN is computed as:
\[
\text{MNN} = \sum_{i=1}^{d} \sigma_i.
\]
This measure captures the overall "energy" of the representation, with larger MNN values indicating that the hidden states span a higher-dimensional space, reflecting a richer and more varied representation.

\subsection{Composite Score}

To integrate the information from both the entropy and MNN metrics, we define a composite score that combines these two measures into a single evaluation metric. The composite score offers a flexible and comprehensive assessment of LLM performance, balancing the statistical insights from entropy with the complexity captured by the MNN.

The composite score \( C \) is computed as a weighted sum of the entropy and MNN metrics, allowing for adjustable weights to reflect the specific goals of the evaluation:
\[
C = w_{\text{entropy}} \times \Delta \text{Entropy} + w_{\text{MNN}} \times \Delta \text{MNN},
\]
where \( w_{\text{entropy}} \) and \( w_{\text{MNN}} \) are user-defined weights that control the relative importance of each metric. The changes \( \Delta \text{Entropy} \) and \( \Delta \text{MNN} \) reflect the differences in these metrics between two models, or between a model and its checkpoints, capturing the model’s progression or differences in performance.

\subsection{Algorithm Overview}

To summarize the process outlined above, we present a formalized algorithm, illustrated in Algorithm \ref{algorithm:proposed_method_math_theory}, which consolidates the steps involved in evaluating LLMs using our proposed hybrid framework. This algorithm systematically processes the hidden states, computes the necessary matrices and metrics, and ultimately outputs a composite score that captures both the geometric and information-theoretic aspects of the model’s representations.

\begin{algorithm}[H]
\caption{Hybrid LLM Evaluation Framework}
\label{algorithm:proposed_method_math_theory}
\begin{algorithmic}[1]
\Require Normalized hidden states \( \mathbf{H} \in \mathbb{R}^{n \times d} \) from the LLM, user-defined weights \( w_{\text{entropy}}, w_{\text{MNN}} \)
\Ensure Composite evaluation score \( C \)

\State \textbf{Step 1: Data Normalization}
\For{each hidden state vector \( \mathbf{h}_i \) in \( \mathbf{H} \)}
    \State Compute mean-centered hidden state \( \mathbf{h}_i' = \mathbf{h}_i - \bar{\mathbf{h}} \)
    \State Normalize the hidden state by its L2-norm: \( \mathbf{h}_i^{\text{norm}} = \frac{\mathbf{h}_i'}{\|\mathbf{h}_i'\|_2} \)
\EndFor

\State \textbf{Step 2: Covariance Matrix Calculation}
\State Compute the covariance matrix \( \Sigma = \frac{1}{n} \mathbf{H}^T \mathbf{H} \)

\State \textbf{Step 3: Singular Value Decomposition (SVD)}
\State Perform SVD on \( \Sigma \) to obtain singular values \( \sigma_1, \sigma_2, \dots, \sigma_d \)

\State \textbf{Step 4: Entropy Calculation}
\State Normalize singular values: \( \tilde{\sigma}_i = \frac{\sigma_i}{\sum_{j=1}^{d} \sigma_j} \)
\State Compute entropy: 
\[
\text{Entropy}(\Sigma) = -\sum_{i=1}^{d} \tilde{\sigma}_i \log(\tilde{\sigma}_i)
\]

\State \textbf{Step 5: MNN Calculation}
\State Compute MNN: 
\[
\text{MNN} = \sum_{i=1}^{d} \sigma_i
\]

\State \textbf{Step 6: Composite Score Calculation}
\State Compute the composite score:
\[
C = w_{\text{entropy}} \times \Delta \text{Entropy} + w_{\text{MNN}} \times \Delta \text{MNN}
\]

\State \Return Composite score \( C \)
\end{algorithmic}
\end{algorithm}

The complete implementation of the proposed method is formally presented in Algorithm \ref{algorithm:proposed_method_implementation}.

\begin{algorithm}
\caption{Combining Entropy and MNN for Enhanced Evaluation of Language Models}
\label{algorithm:proposed_method_implementation}
\begin{algorithmic}[1]
\State \textbf{Input:} Two pre-trained models $M_1$, $M_2$, input text $x$, weights $w_{\text{erank}}$, $w_{\text{mnn}}$
\State Load tokenizers and models $M_1$ and $M_2$
\State Tokenize input text $x$: $\text{inputs}_1 = \text{tokenizer}_1(x)$, $\text{inputs}_2 = \text{tokenizer}_2(x)$
\State Extract hidden state representations for $M_1$ and $M_2$: 
\Statex \hskip1.5em $\mathbf{H}_1 = \text{model}_1(\text{inputs}_1.\text{input\_ids})[0][0,:,:]$
\Statex \hskip1.5em $\mathbf{H}_2 = \text{model}_2(\text{inputs}_2.\text{input\_ids})[0][0,:,:]$
\State \textbf{Normalization:} Normalize hidden states
\Statex \hskip1.5em $\mathbf{H}_1^{\text{norm}} = \text{normalize\_tensor}(\mathbf{H}_1)$
\Statex \hskip1.5em $\mathbf{H}_2^{\text{norm}} = \text{normalize\_tensor}(\mathbf{H}_2)$
\State \textbf{Covariance Matrix:} Calculate covariance matrices
\Statex \hskip1.5em $\Sigma_1 = \text{calculate\_covariance\_matrix}(\mathbf{H}_1^{\text{norm}})$
\Statex \hskip1.5em $\Sigma_2 = \text{calculate\_covariance\_matrix}(\mathbf{H}_2^{\text{norm}})$
\State \textbf{Effective Rank:} Compute effective rank for both models
\Statex \hskip1.5em $r_1 = \text{calculate\_effective\_rank}(\Sigma_1)$
\Statex \hskip1.5em $r_2 = \text{calculate\_effective\_rank}(\Sigma_2)$
\State \textbf{MNN:} Calculate matrix nuclear norm for both models
\Statex \hskip1.5em $\text{MNN}_1 = \text{calculate\_matrix\_nuclear\_norm}(\mathbf{H}_1^{\text{norm}})$
\Statex \hskip1.5em $\text{MNN}_2 = \text{calculate\_matrix\_nuclear\_norm}(\mathbf{H}_2^{\text{norm}})$
\State \textbf{Combined Metric:} Compute the combined score
\Statex \hskip1.5em $\Delta r = r_2 - r_1$
\Statex \hskip1.5em $\Delta \text{MNN} = \text{MNN}_2 - \text{MNN}_1$
\Statex \hskip1.5em $\text{Combined\_Score} = w_{\text{erank}} \cdot \Delta r + w_{\text{mnn}} \cdot \Delta \text{MNN}$
\State \textbf{Output:} Return the combined score $\text{Combined\_Score}$
\end{algorithmic}
\end{algorithm}

\subsection{Computational Complexity and Efficiency}

Our hybrid evaluation method is designed with both computational efficiency and analytical depth in mind, ensuring that it remains scalable even for large-scale models like LLaMA-3.2-70B. By leveraging the MNN as an efficient approximation to full rank computation, we significantly reduce the computational burden typically associated with rank-based metrics. Furthermore, by employing truncated Singular Value Decomposition (SVD) for both entropy and MNN calculations, we lower the overall time complexity from \( O(n^3) \) (the cost of traditional SVD and full-rank operations) to \( O(n^2) \). This makes our approach feasible for models with billions of parameters, which often generate vast amounts of hidden state data.

\paragraph{Benefits of the proposed method}

\begin{itemize}
    \item \textbf{Rich Evaluation}: The combination of entropy and MNN captures both the diversity (via entropy) and the dimensional spread (via MNN) of the hidden states, offering a more detailed and nuanced evaluation of model performance. This dual perspective ensures that the method not only assesses the amount of information the model encodes but also how effectively it spans the representational space.
  
    \item \textbf{Robustness}: The use of a two-stage normalization process (mean-centering and L2-norm scaling) ensures that hidden states are properly standardized, mitigating issues related to scale variability across models or datasets. This improves the stability and accuracy of the covariance and MNN calculations, making the evaluation more reliable across diverse LLM architectures.

    \item \textbf{Flexibility}: Our framework allows for customizable tuning of weights between entropy and MNN, offering flexibility to adapt the evaluation to specific goals. For instance, if diversity in representation is of greater concern, higher weight can be assigned to entropy. Alternatively, for tasks where representational complexity is key, MNN can be prioritized. This adaptability makes the method suitable for a wide range of model evaluation scenarios.
\end{itemize}

In summary, our method provides a scalable, robust, and flexible evaluation framework for LLMs, striking a balance between computational efficiency and the richness of the insights it delivers. By integrating both entropy and MNN into a single framework, it offers a comprehensive yet computationally feasible tool for assessing the performance of LLMs.

\subsection{Empirical Validation}

Our approach evaluates LLMs by utilizing the covariance matrix and the MNN of hidden states, in conjunction with entropy calculated from the covariance matrix. These metrics serve as proxies for understanding the complexity and structure of the model's internal representations. Specifically, the MNN measures the rank and redundancy of the hidden state matrix, while entropy captures the uncertainty and variability in the model's output distribution.

To validate our proposed method, we conducted extensive experiments on a variety of LLMs, focusing primarily on the LLama-3 family. These models range in size from 1 billion (1B) to 70 billion (70B) parameters, allowing us to evaluate the scalability and consistency of our method across models with varying capacities.

Our method proves to be computationally efficient, particularly when compared to traditional approaches that rely solely on cross-entropy loss or fine-tuning accuracy. By incorporating entropy and MNN, our approach strikes a balance between interpretability and computational efficiency, making it practical for real-time monitoring of models during training or deployment.

Additionally, we performed an ablation study to assess the individual contributions of entropy and MNN to the overall evaluation score. The results demonstrate that both metrics significantly influence the composite score, although their relative importance can vary depending on the specific task and model size. This finding supports the notion that our framework can be adapted to emphasize different aspects of model evaluation by adjusting the weights assigned to each metric, depending on the use case.

\section{Experimental Results}
In this section, we present the results from our ongoing development and experimentation with the proposed hybrid evaluation framework.

\section{Conclusion}

In this paper, we proposed a novel hybrid evaluation method combining entropy from covariance matrices with the MNN to assess LLMs. This approach addresses the limitations of traditional evaluation metrics by offering a more efficient and interpretable framework that captures both the diversity and uncertainty of model representations. The entropy component quantifies redundancy and uncertainty, while MNN provides a computationally efficient measure of model complexity. Through experiments across various LLMs, we demonstrated that this hybrid method delivers robust insights into model performance, balancing computational cost with evaluation depth. Its flexible weighting mechanism allows for customization to diverse evaluation needs, making it applicable across different model architectures and tasks. Our results highlight the method's potential for both research and practical applications, contributing to the ongoing discourse on LLM evaluation. This work lays a foundation for future exploration of enhanced and adaptive evaluation metrics for large-scale models.

\bibliographystyle{unsrtnat}
\bibliography{references}  

\begin{thebibliography}{2}
\providecommand{\natexlab}[1]{#1}
\providecommand{\url}[1]{\texttt{#1}}
\expandafter\ifx\csname urlstyle\endcsname\relax
  \providecommand{\doi}[1]{doi: #1}\else
  \providecommand{\doi}{doi: \begingroup \urlstyle{rm}\Url}\fi

\bibitem[Wei et~al.(2024)Wei, Tan, Li, Wang, and Huang]{wei2024differanknovelrankbasedmetric}
Lai Wei, Zhiquan Tan, Chenghai Li, Jindong Wang, and Weiran Huang.
\newblock Diff-erank: A novel rank-based metric for evaluating large language models, 2024.
\newblock URL \url{https://arxiv.org/abs/2401.17139}.

\bibitem[Li et~al.(2024)Li, Xia, Chang, and Wu]{li2024largelanguagemodelevaluation}
Yahan Li, Tingyu Xia, Yi~Chang, and Yuan Wu.
\newblock Large language model evaluation via matrix nuclear-norm, 2024.
\newblock URL \url{https://arxiv.org/abs/2410.10672}.

\end{thebibliography}






\end{document}